\documentclass[10pt,twocolumn,letterpaper]{article}

\usepackage{wacv}
\usepackage{times}
\usepackage{epsfig}
\usepackage{graphicx}
\usepackage{amsmath}
\usepackage{amssymb}
\usepackage{algorithmic}
\usepackage{graphicx}
\usepackage{textcomp}
\usepackage{longtable}
\usepackage{subcaption}
\usepackage[table,xcdraw]{xcolor}
 \usepackage{dblfloatfix}
\usepackage{float}
 \usepackage{booktabs}
\usepackage{bigstrut}
\usepackage{multirow}
\usepackage{color}
\usepackage{caption}
\usepackage{fancyhdr}
\usepackage{listings} 
\usepackage{changepage}
\usepackage{pgfplots}
\usepackage{balance}
\usepackage{enumitem}
\usepackage{pifont}
\newcommand{\cmark}{\ding{51}}%
\usepackage[accsupp]{axessibility}
\usepackage[pagebackref=true,breaklinks=true,letterpaper=true,colorlinks,bookmarks=false]{hyperref}

\def\wacvPaperID{0358} 

\wacvapplicationstrack 
\wacvfinalcopy 

\ifwacvfinal
\usepackage[breaklinks=true,bookmarks=false]{hyperref}
\else
\usepackage[pagebackref=true,breaklinks=true,colorlinks,bookmarks=false]{hyperref}
\fi

\begin{document}

\title{Fingervein Verification using Convolutional Multi-Head Attention Network}

\author{Raghavendra Ramachandra\\
Norwegian University of Science and Technology (NTNU)\\
Norway\\
{\tt\small Email: raghavendra.ramachandra@ntnu.no}
\and
Sushma Venkatesh\\
AiBA AS\\
Norway\\
{\tt\small Email: sushma@aiba.ai}
}

\maketitle
\thispagestyle{empty}

\begin{abstract}

Biometric verification systems are deployed in various security-based access-control applications that require user-friendly and reliable person verification. Among the different biometric characteristics, fingervein biometrics have been extensively studied owing to their reliable verification performance. Furthermore, fingervein patterns reside inside the skin and are not visible outside; therefore, they possess inherent resistance to presentation attacks and degradation due to external factors.  In this paper, we introduce a novel fingervein verification technique using a convolutional multihead attention network called VeinAtnNet. The proposed VeinAtnNet is designed to achieve light weight with a smaller number of learnable parameters while extracting discriminant information from both   normal and enhanced fingervein images.  The proposed VeinAtnNet was trained on the newly constructed fingervein dataset with 300 unique fingervein patterns that were captured in multiple sessions to obtain 92 samples per unique fingervein. Extensive experiments were performed on the newly collected dataset FV-300 and the publicly available FV-USM and FV-PolyU fingervein dataset. The performance of the proposed method was compared with five state-of-the-art fingervein verification systems, indicating the efficacy of the proposed VeinAtnNet. 

\end{abstract}

\section{Introduction}

\begin{figure}[t!]
\begin{center}
\includegraphics[width=1.0\linewidth]{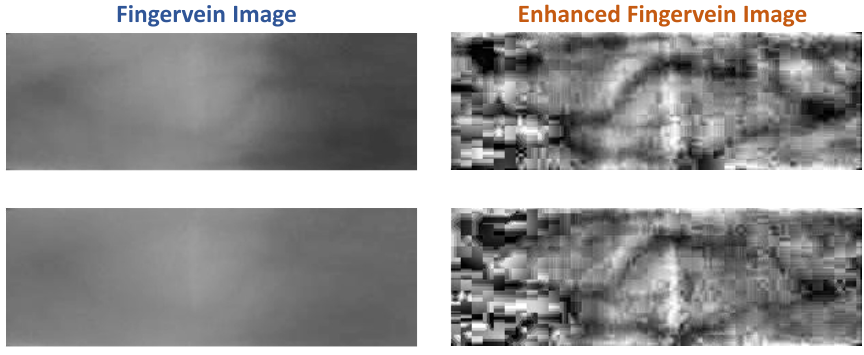}
\end{center}
   \caption{Example fingervein images with and without image enhancement for the same identity collected in first (top row) and second session (bottom row).}
\label{fig:intro}
\end{figure}

Biometric verification systems have enabled magnitude of access control applications including border control,  smartphone access, banking, and finance applications. Fingervein biometric characteristics are widely deployed in various applications, particularly in banking sector. Fingervein biometrics represent the vein structure underneath the skin of the finger, which can be captured using near-infrared sensing. The blood flow in the fingervein absorbs near-infrared light and appears dark compared to the neighborhood region, indicating the visibility of the fingervein (refer Figure \ref{fig:intro}). The fingervein structure has been shown to be unique \cite{Hitachi, shahin2007FVUniqueness, Dorsal_Rag} between fingers of same data subject and between the data subjects. Compared to other biometric characteristics, fingervein biometrics are known for their accuracy and usefulness, and are less vulnerable to distortion. Furthermore, fingervein biometrics provide a natural way of protecting biometric features, as they reside inside the skin and thus more challenging to spoof.

Fingervein biometrics have been widely studied in the literature, resulting in various fingervein biometric verification algorithms \cite{FV_Review, FV_REview2}.  Early works are based on extracting fingervein patterns such that the vein region is labeled as one and the background is labeled as zero. Techniques such as Maximum Curvature Points (MCP) \cite{miura2007extraction}, Repeated Line Tracking (RLT) \cite{RepeatedlineTracking}, Wide Line Detectors (WLD) \cite{FVPen1}, Mean Curvature (MC) \cite{song2011fingerMC} and Randon transform \cite{qin2017Randon} have been developed for reliable fingervein recognition.  As these techniques can extract the structure of the fingervein pattern, the use of a simple comparator based on template matching using correction can achieve reliable performance. However, these features are sensitive to a small degree of fingervein rotation, noise, and reflection properties of the skin and NIR illuminator.

\begin{table*}[htp]
\centering
\caption{State-of-the-art fingervein verification using  deep learning techniques}
\label{tab:SOTAtable}
\resizebox{1.8\columnwidth}{!}{%
\begin{tabular}{|l|l|l|}
\hline
\rowcolor[HTML]{CBCEFB} 
\textbf{Authors} &
  \textbf{Year} &
  \textbf{Deep Learning   Technique} \\ \hline \hline
Huafeng Qin et al., \cite{qin2015finger} &
  2015 &
  Serial CNN architecture   with 3 convolution layers and 2 fully connected layer. \\ \hline
Itqan et al., \cite{itqan2016user} &
  2016 &
  Serial CNN architecture   with 3 convolution layers and 1  fully   connected layer. \\ \hline
Syafeeza Radzi et al., \cite{radzi2016finger} &
  2016 &
  Serial CNN architecture   with 2 convolution layers. \\ \hline
Huafeng Qin et al., \cite{qin2017deep} &
  2017 &
  Serial CNN architecture   with 4 convolution layers. Path based training of CNN. \\ \hline
Cihui Xie et al., \cite{xie2019finger} &
  2019 &
  Siamese network   with 5 conventional layers and triplet loss function. \\ \hline
\begin{tabular}[c]{@{}l@{}}Jong Min Song et al.,\cite{song2019finger}\end{tabular} &
  2019 &
  \begin{tabular}[c]{@{}l@{}}Serial CNN   architecture with 8 convolution layers. Composite fingervein image is   \\ generated by converting the 1-channel input image to 3-channel input image.\end{tabular} \\ \hline
Rig Das et al., \cite{das2018convolutional} &
  2018 &
  Serial CNN   architecture with 5 convolution layers. \\ \hline
\begin{tabular}[c]{@{}l@{}}Hyung Gil Hong et al.,\cite{ hong2017convolutional}\end{tabular} &
  2017 &
  Serial CNN   architecture with 12 convolution layers and 3 fully connected layers. \\ \hline
Su Tnag et al., \cite{tang2019finger} &
  2019 &
  Siamese network   with residual CNN architecture. \\ \hline
Borui Hou et   al., \cite{hou2019convolutional} &
  2019 &
  Convolutional   autoencoder. \\ \hline
Junying Zeng et al., \cite{zeng2020finger} &
  2020 &
  Deformable   convolution with U-NET type architecture. \\ \hline
Ridvan Salih   Kuzu et al., \cite{kuzu2020fly} &
  2020 &
  \begin{tabular}[c]{@{}l@{}}Serial CNN   architecture with 6 convolution layers and 2 fully connected layers \\ with LSTM   for classification.\end{tabular} \\ \hline
Hengyi Ren et al., \cite{ren2021finger} &
  2021 &
  \begin{tabular}[c]{@{}l@{}}Feature extraction   using ResNet with squeeze and excitation \\ on the encrypted fingervein images.\end{tabular} \\ \hline
R$\iota$dvan   Salih Kuzu et al., \cite{kuzu2021loss} &
  2021 &
  \begin{tabular}[c]{@{}l@{}}Custom DenseNet   161 with additive angular penalty and \\ large margin cosine penalty loss   function.\end{tabular} \\ \hline
Weili Yang et al., \cite{yang2022multi} &
  2022 &
  Multi-view   fingervein with individual CNNs and view pooling. \\ \hline
Huafeng Qin et al., \cite{yang2022multi} &
  2022 &
  U-Net based   architecture with attention module. \\ \hline
Tingting Chai   et al., \cite{chai2022shape} &
  2022 &
  Serial CNN   architecture with 5 convolution layers and one fully connected layer. \\ \hline
Ismail et   al., \cite{boucherit2022fingerFUSION} &
  2022 &
  Serial CNN   architecture with 3 convolution layers and two fully connected layer. \\ \hline
Weiye Liu et   al., \cite{liu2023mmran} &
  2023 &
  Residual   Attention block with inception architecture. \\ \hline
Zhongxia Zhang et al., \cite{zhang2023finger} &
  2023 &
  Light weight   CNN with spatial and channel attention module. \\ \hline
Chunxin Fang et al., \cite{fang2023finger} &
  2023 &
  Light weight Siamese   network with attention module. \\ \hline
Bin Wa et al., \cite{ma2023fingerMultiAtten} &
  2023 &
  \begin{tabular}[c]{@{}l@{}}Serial CNN   architecture with 3 convolution layers and\\  bilinear pooling with multiple   attention module.\end{tabular} \\ \hline\hline
\rowcolor[HTML]{EFEFEF} 
\textbf{This work} &
  \textbf{2024} &
  \begin{tabular}[c]{@{}l@{}}\textbf{Serial CNN architecture   with 3 convolution layers and} \\ \textbf{multi-head attention module connected in parallel}   \\ \textbf{with normal and enhanced fingervein.}\end{tabular} \\ \hline \hline
\end{tabular}%
}
\end{table*}


The global feature representation of fingervein patterns such as Local Binary Patterns (LBP) \cite{FV_LBP}, Gabor filters \cite{FV_HoG}, Local Directional Code \cite{sikarwar2016fingerLDA}, Wavelet Transform \cite{park2011finger}, Histogram of Gradients (HoG)  \cite{FV_HoG} and pyramid image features \cite{lu2018pyramid} are also developed for the fingervein verification. These features are often used with Support Vector Machines (SVM) or Euclidean distances as comparators. As these techniques are based on global features, they are highly sensitive to variations in finger rotation and illumination.

The representation of a fingervein image to binary codes was developed to improve template security, together with reliable verification.  Binary coding techniques include Discriminative Binary Codes \cite{liu2017discriminative}, binary hash codes \cite{su2019learning}, DoG code \cite{Dorsal_Rag}, ordinal code \cite{Dorsal_Rag}, contour Code \cite{Dorsal_Rag} and competitive codes \cite{Dorsal_Rag}. Because these techniques can generate binary codes for the finger vein, the Hamming distance is used as the comparator. Binary coding techniques exhibit good verification accuracy; however, these features are sensitive to variations in rotation and illumination.

\begin{figure*}[htp]
\begin{center}
\includegraphics[width=1.0\linewidth]{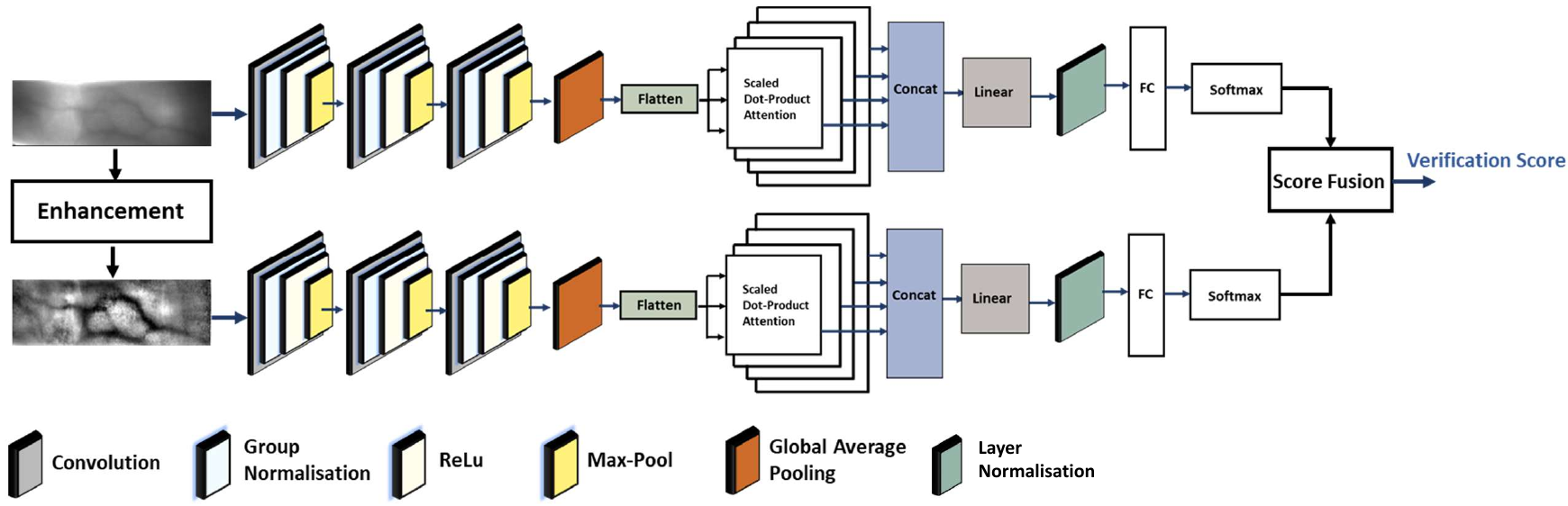}
\end{center}
   \caption{Block diagram of the proposed method for fingervein verification}
\label{fig:Pro}
\end{figure*}

Deep-learning-based fingervein verification has been extensively studied in the literature. Table \ref{tab:SOTAtable} summarizes the deep-learning-based techniques proposed for fingervein recognition. Early works are based on the serial convolution architecture, which is inspired from existing CNN architectures that are evaluated on the ImageNet dataset.  Both shallow serial CNN networks with two convolution layers and a deep CNN network with 12 convolutional layers have been studied in the literature. However, the quantitative results indicate that lightweight serial networks with a smaller number of convolution layers exhibit better performance than deep serial networks. The possible degraded performance of deep serial networks can be attributed to limited data availability. The use of a pre-trained CNN for feature extraction has also been explored in the literature, together with fine-tuning and augmented pre-trained CNN networks. The quantitative results reported indicate a performance similar to that of end-to-end trained deep CNN networks. The Siamese network for fingervein verification was studied using different CNN configurations and U-NET-based architectures. The quantitative performance is similar to that of the serial CNN architecture. Recently, attention modules with lightweight (three to four convolution layers) serial networks have been widely explored. Different types of attention modules, including spatial, channel, and multi-attention modules, were introduced. The quantitative performance of the attention networks are  comparable to that of other deep learning-based techniques implemented for fingervein verification.

Even though the deep learning techniques are widely studied for the reliable fingervein verification, the existing deep learning techniques indicates the following drawbacks: (a) limited data: Existing techniques are evaluated on the small-scale datasets that has 6-12 samples per data subject. This limits the effectiveness of deep learning and leads to an over fitting. (b) Lack of a consistent evaluation protocol: Even though most of the existing works have used public datasets, the evaluation protocols are not consistent across existing studies. This results in a limited comparison of existing techniques for finger vein verification.  In this study, we address the above-mentioned limitation by introducing a new large-scale dataset with 75 data subjects, resulting in 300 unique identities (as we collected four fingers per data subject). For each unique fingervein, we collected 92 samples in multiple sessions, varying from 1-4 days duration. Furthermore, we propose a novel lightweight CNN architecture based on a convolutional multi-head attention module. The main contributions of this study are as follows:
\begin{itemize} [leftmargin=*,noitemsep, topsep=0pt,parsep=0pt,partopsep=0pt]
\item 	 A novel fingervein verification technique based on a convolutional multi-head attention network (VeinAtnNet) is proposed. 
\item 	Introduced a new fingervein dataset with 300 unique identities captured from 75 data subjects, resulting in $300 \times 92 = 27600$ fingervein images. The dataset is available publicly for research purpose. 
\item 	Extensive experiments were performed on both the newly introduced dataset and the publicly available FV-USM and FV-PolyU datasets. The performance of the proposed method was compared with that of five state-of-the-art fingervein verification methods. 
\end{itemize}

The rest of the paper is organised as follows: Section \ref{sec:pro} discuss the proposed method for the fingervein verification, Section \ref{sec:exp} presents the quantitative results of the proposed method with the state-of-the-art techniques and Section \ref{sec:conc}  draws the conclusion. 
\section{Proposed Method}
\label{sec:pro}

Figure \ref{fig:Pro} shows a block diagram of the proposed VeinAtnNet architecture for reliable fingervein verification. The novelty of the proposed approach is that it leverages the convolutional Multi-Head Attention (MHA) framework to achieve accurate and reliable fingervein verification.  The utility of MHA, together with convolutional features, leads to a discriminant feature representation that can contribute to the robust  performance of the fingervein verification.

The proposed VeinAtnNet is a lightweight architecture with three Consecutive Convolution Layers (CCL) and a Multihead Self-Attention (MSA) mechanism.  VeinAtnNet is connected independently with normal and enhanced fingervein images whose comparison scores from the softmax layer are fused to make the final verification decision.  Given the captured fingervein image, preprocessing is performed using Contrast Limited Adaptive Histogram Equalization (CLAHE) \cite{FVPen5} to enhance the fingervein pattern. In this work, we employed the Contrast Limited Adaptive Histogram (CLAHE) as the fingervein enhancement method by considering (a) the high quality of the fingervein enhancement achieved when compared to other enhancement techniques, as discussed in \cite{FV_REview2}. (b) Widely employed enhancement techniques in fingervein literature that have reported high verification accuracy. 
Both the normal (without enhancement) and enhanced fingervein images were resized to $224 \times 224 \times 3$ pixels.  The CCL performs the initial feature learning of the fingervein images, which is further processed to obtain a rich feature representation using MSA. Given the fingervein image $F_{v}^{R \times C \times D}$, the final output features of MSA can  be represented as follows: 
\begin{equation}
F_{MSA} = MSA(F_{CCL}); where  F_{CCL} = CCL(F_{v});
\end{equation}
Where, $F_{MSA}$ denote the output features from MSA, $F_{CCL}$ denotes the output features of CCL block. The $F_{MSA}$ is then used with softmax classifier to make the final decision. In the following, we discuss the building blocks of the proposed VeinAtnNet. 

\subsection{Consecutive Convolution Layers (CCL)}
The CCL has three convolution modules that are serially connected. Each convolution module has four different convolution layers (conv): a group normalization (norm), an activation function layer (ReLu), and a pooling layer (maxpool). Three convolution layers were used to extract the global features from the fingervein images. The conv-1 layer has a filter size of $7 \times 7$ the conv-2 layer has a $5 \times 5$, conv-3 has a $3 \times 3$ filter, and the number of filters in all three conv layers is set to 32. The gradual decrease in filter size ensures fine grinding (from global to local) of the fingervein features. The convolution features were normalized using group normalization, which reduced the sensitivity of the network for initialization.   In particular, we employed group normalization because it outperforms batch normalization with a small size. The normalized features are then fed to the activation unit (ReLU), which can introduce sparsity and improve the network training speed. Finally, a pooling operation was performed to achieve a compact feature representation. In this study, we employed max pooling, which can capture texture information suitable for fingervein verification. The output after three convolution modules is then passed through the group-average pooling layer to obtain a compact representation of the features. Finally, the features were flattened before being fed into the MSA module.

\subsection{Multihead Self-Attention (MSA)}
The features from the CCL module are then fed to the MSA module to further refine the features $F_{CCL}$ to extract discriminant features suitable for fingervein verification. In this study, we employed multihead attention \cite{vaswani2017attention} with four different heads and 64 channels for keys and queries. Basically, MSA runs the attention mechanism across all heads multiple times in parallel. The independent attention outputs are then concatenated and transformed linearly. MSA can be represented as follows \cite{vaswani2017attention}:
\begin{equation}
Mu-Head(Q.K,V) = [H_{1}, H_{2}, H_{3}, H_{4}] W
\end{equation}
where $W$ is the learnable parameter and $Q$, $K$ and $V$ represent the queries, keys, and values, respectively.  In this study, we employed scaled dot-product attention across heads using $Q$, $K$ and $V$ as follows:
\begin{equation}
Attention(Q.K,V) = softmax(\frac{QK^{T}}{sqrt(d_{k})})V
\end{equation}
The outcome of the MSA module was passed through the layer normalization layer to generalize the final features.  Finally, the normalized features are passed through the fully connected and softmax layers to obtain the comparison score. 
\subsection{Score Level Fusion}
The proposed VeinAtnNet was employed independently on normal and enhanced finger vein images. Thus, given the test fingervein image, the proposed method provides two comparison scores corresponding to normal and enhanced fingerveins. We combined these two comparison scores using the sum rule to make the final verification decision. 
Let the comparison score from the normal fingervein image be $C_{n}$ and enhanced fingervein image be $C_{e}$, then final verification score is computed as $V_{s} = (C_{n} + C_{e})$.

\subsection{Implementation  Details}
The proposed network is based on Adaptive Moment Estimation (ADAM) optimization to calculate loss. In this work, we employ the cross-entropy loss, which can be defined as $-\frac{1}{N} \sum_{n=1}^{N} \sum_{i=1}^{K} (T_{ni} \log(Y_{ni})) + (1-T_{ni})\log(1-Y_{ni}) $, where $N$ and $K$ denote the number of samples and classes, respectively, $T_{ni}$ is the corresponding target value to $Y_{ni}$. During training, the learning rate was set to 0.0001, the mini-batch size was set to 16, and the number of epochs was set to 150. Furthermore, we performed data augmentation, which included image reflection, translation, rotation, reflection, scaling, and random noise with three different variances. This resulted in nine different images for every image used in training the proposed method. Finally, the proposed method is lightweight with only 58.2 K learnable parameters. While the existing SOTA employed in this work namely; Bin Wa   et al., \cite{ma2023fingerMultiAtten} has approximately 17.8M and Ismail et al., \cite{boucherit2022fingerFUSION} has approximately  467.1K learnable parameters respectively.  

\section{Experiments and Results}
\label{sec:exp}

In this section, we discuss the quantitative results of the proposed and existing  fingervein verification algorithms. The quantitative performance is presented using the False Match Rate (FMR) and False Non-Match Rate (FNMR), together with the Equal Error Rate (EER) value computed  at FMR = FNMR.  The performance of the proposed method was compared with recently proposed fingervein recognition algorithms based on multiple attentions \cite{ma2023fingerMultiAtten} and deep fusion \cite{boucherit2022fingerFUSION} by considering their verification performance. Furthermore, we compared the performance of the proposed method with well-established fingervein verification techniques, such as MCP \cite{miura2007extraction}, RLT\cite{RepeatedlineTracking} and WLD \cite{FVPen1}. In the following section, we describe the newly collected fingervein dataset, followed by the quantitative results.

\subsection{FV-300 Fingervein dataset}
\begin{figure}[htp]
\begin{center}
\includegraphics[width=1.0\linewidth]{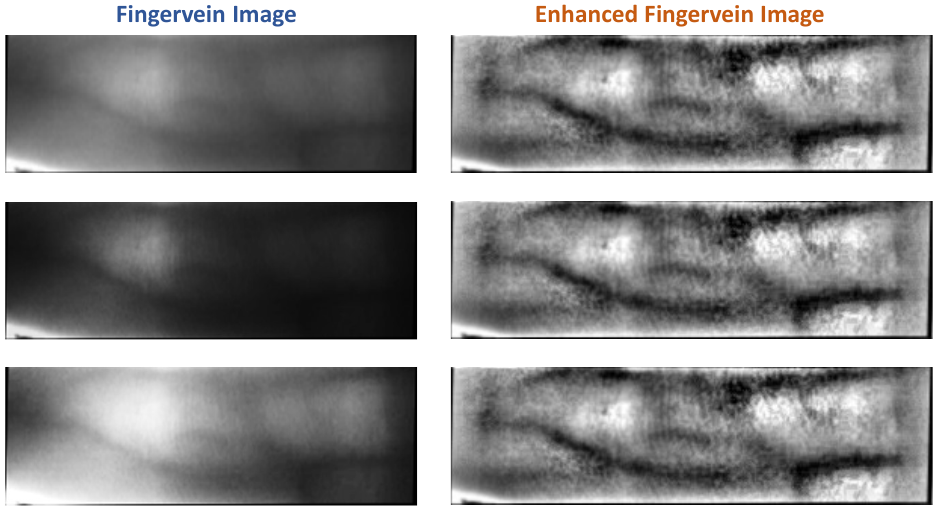}
\end{center}
   \caption{Example fingervein images from FV-300 dataset representing same identity captured in three different sessions.}
\label{fig:fv300}
\end{figure}

In this study, we introduced a new fingervein dataset comprising 300 unique fingerveins corresponding to  75 unique data subjects. The fingervein images were collected using a custom camera system desired using a monochrome CMOS camera with a resolution of $744 \times 480$ pixels with two lighting sources to illuminate the finger from both the back and side. The design aspects of the finger vein capture device were inspired by \cite{8672597}. The data collection was carried out under indoor conditions, and for every data subject, two fingers (index and middle) were captured from both the left and right hands, resulting in four unique fingers. For each data subject, we captured 92 fingervein images corresponding to individual fingers in multiple sessions. The duration between sessions varies from to 1-4 days. The FV-300 dataset contained 75 data subjects $\times$ 4 fingers $\times$ 92 = 27600 fingervein samples. Figure \ref{fig:fv300} shows an example of the fingervein images from the FV-300 dataset. 

\begin{table*}[htp]
\centering
\caption{Quantitative Performance of the proposed and state-of-the-art fingervein verification methods }
\label{tab:Results}
\resizebox{1.3\columnwidth}{!}{%
\begin{tabular}{|l|l|l|lll|}
\hline
 &
   &
   &
  \multicolumn{3}{l|}{\textbf{TAR = (100-FNMR\%) @ FMR =}} \\ \cline{4-6} 
\multirow{-2}{*}{\textbf{Data set}} &
  \multirow{-2}{*}{\textbf{Algorithms}} &
  \multirow{-2}{*}{\textbf{EER(\%)}} &
  \multicolumn{1}{l|}{\textbf{1\%}} &
  \multicolumn{1}{l|}{\textbf{0.1\%}} &
  \textbf{0.01\%} \\ \hline
\cellcolor[HTML]{EFEFEF} &
  MCP \cite{miura2007extraction} &
  4.74 &
  \multicolumn{1}{l|}{88.93} &
  \multicolumn{1}{l|}{64.41} &
  44.35 \\ \cline{2-6} 
\cellcolor[HTML]{EFEFEF} &
  RLT \cite{RepeatedlineTracking} &
  31.30 &
  \multicolumn{1}{l|}{19.35} &
  \multicolumn{1}{l|}{9.52} &
  4.68 \\ \cline{2-6} 
\cellcolor[HTML]{EFEFEF} &
  WLD \cite{FVPen1} &
  13.55 &
  \multicolumn{1}{l|}{77.47} &
  \multicolumn{1}{l|}{74.11} &
  73.15 \\ \cline{2-6} 
\cellcolor[HTML]{EFEFEF} &
  Ismail et al., \cite{boucherit2022fingerFUSION}&
  9.14 &
  \multicolumn{1}{l|}{72.92} &
  \multicolumn{1}{l|}{40.95} &
  7.34 \\ \cline{2-6} 
\cellcolor[HTML]{EFEFEF} &
  Bin Wa   et al., \cite{ma2023fingerMultiAtten} &
  19.94 &
  \multicolumn{1}{l|}{21.25} &
  \multicolumn{1}{l|}{5.31} &
  0.95 \\ \cline{2-6} 
\multirow{-6}{*}{\cellcolor[HTML]{EFEFEF}FV300} &
  \textbf{Proposed Method} &
  \textbf{0.54} &
  \multicolumn{1}{l|}{\textbf{99.73}} &
  \multicolumn{1}{l|}{\textbf{99.13}} &
  \textbf{90.36} \\ \hline \hline
\cellcolor[HTML]{ECF4FF} &
  MCP \cite{miura2007extraction}&
  17.74 &
  \multicolumn{1}{l|}{46.95} &
  \multicolumn{1}{l|}{27.25} &
  18.36 \\ \cline{2-6} 
\cellcolor[HTML]{ECF4FF} &
  RLT \cite{RepeatedlineTracking} &
  29.63 &
  \multicolumn{1}{l|}{32.29} &
  \multicolumn{1}{l|}{29.70} &
  29.16 \\ \cline{2-6} 
\cellcolor[HTML]{ECF4FF} &
  WLD \cite{FVPen1} &
  18.17 &
  \multicolumn{1}{l|}{54.17} &
  \multicolumn{1}{l|}{34.89} &
  16.73 \\ \cline{2-6} 
\cellcolor[HTML]{ECF4FF} &
  Ismail et al., \cite{boucherit2022fingerFUSION}&
  41.78 &
  \multicolumn{1}{l|}{4.38} &
  \multicolumn{1}{l|}{1.16} &
  0.34 \\ \cline{2-6} 
\cellcolor[HTML]{ECF4FF} &
  Bin Wa   et al., \cite{ma2023fingerMultiAtten} &
  37.53 &
  \multicolumn{1}{l|}{5.25} &
  \multicolumn{1}{l|}{1.65} &
  0.35 \\ \cline{2-6} 
\multirow{-6}{*}{\cellcolor[HTML]{ECF4FF}FV-USM} &
  \textbf{Proposed Method} &
  \textbf{15.35} &
  \multicolumn{1}{l|}{\textbf{40.45}} &
  \multicolumn{1}{l|}{\textbf{14.25}} &
  \textbf{5.11} \\ \hline \hline
\cellcolor[HTML]{FFCCC9} &
  MCP \cite{miura2007extraction} &
  14.25 &
  \multicolumn{1}{l|}{52.29} &
  \multicolumn{1}{l|}{34.40} &
  20.64 \\ \cline{2-6} 
\cellcolor[HTML]{FFCCC9} &
  RLT \cite{RepeatedlineTracking}&
  33.48 &
  \multicolumn{1}{l|}{19.26} &
  \multicolumn{1}{l|}{8.25} &
  4.28 \\ \cline{2-6} 
\cellcolor[HTML]{FFCCC9} &
  WLD \cite{FVPen1} &
  16.53 &
  \multicolumn{1}{l|}{65.29} &
  \multicolumn{1}{l|}{48.16} &
  38.53 \\ \cline{2-6} 
\cellcolor[HTML]{FFCCC9} &
  Ismail et al., \cite{boucherit2022fingerFUSION} &
  42.12 &
  \multicolumn{1}{l|}{5.96} &
  \multicolumn{1}{l|}{2.29} &
  1.37 \\ \cline{2-6} 
\cellcolor[HTML]{FFCCC9} &
  Bin Wa   et al., \cite{ma2023fingerMultiAtten}&
  40.90 &
  \multicolumn{1}{l|}{3.21} &
  \multicolumn{1}{l|}{0.45} &
  0.45 \\ \cline{2-6} 
\multirow{-6}{*}{\cellcolor[HTML]{FFCCC9}PolyU} &
  \textbf{Proposed Method} &
  \textbf{5.52} &
  \multicolumn{1}{l|}{\textbf{74.77}} &
  \multicolumn{1}{l|}{\textbf{30.27}} &
  \textbf{19.26} \\ \hline \hline
\end{tabular}%
}
\end{table*}


\begin{figure*}[htp] 
    \centering
    \subfloat[FV-300 Dataset]{%
        \includegraphics[width=0.33\textwidth]{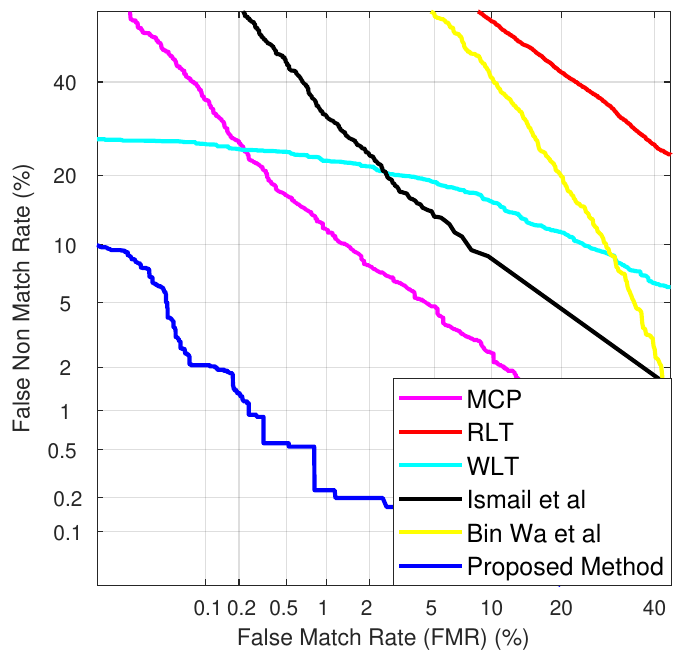}%
        \label{fig:a}%
        }%
    \hfill%
    \subfloat[FV-USM Dataset]{%
        \includegraphics[width=0.33\textwidth]{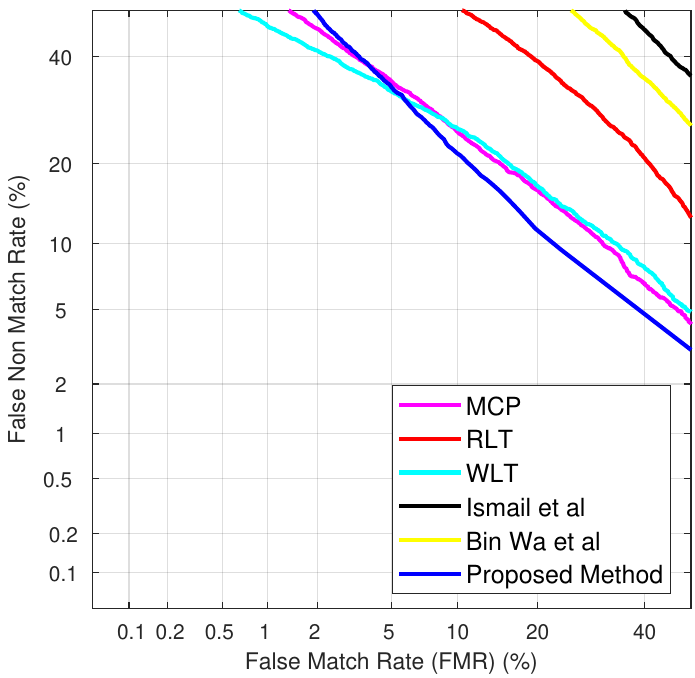}%
        \label{fig:b}%
        }%
         \hfill%
    \subfloat[FV-PolyU Dataset]{%
        \includegraphics[width=0.33\textwidth]{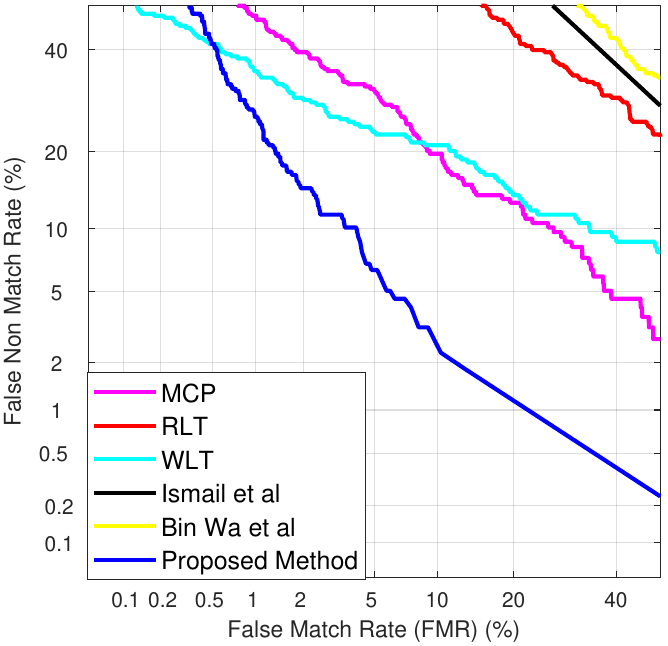}%
        \label{fig:c}%
        }%
    \caption{DET Curves showing the verification performance of the proposed and state-of-the-art fingervein verification methods}
    \label{fig:DET}
\end{figure*}

\subsection{Experimental protocol}
To effectively benchmark the performance of the proposed method, we used three fingervein datasets: FV-300,  FV-USM \cite{asaari2014fusion} and FV-PolyU \cite{FP_FV_Sensor2}. To evaluate the performance of the fingervein algorithm on FV-300 dataset, the fingervein samples corresponding to each finger were divided into three independent sets such that the training set had 70 images, the validation set had 12 images, and the testing set had 10 images.  This resulted in 300 $\times$ 10 = 3000 genuine and 300 $\times$ 299 $\times$  10 = 897000 impostor scores, respectively.

The verification performance of the FV-USM \cite{asaari2014fusion} dataset was evaluated by training the fingervein verification algorithms on FV-300 dataset and fine-tuning the trained networks on the FV-USM dataset. The FV-USM \cite{asaari2014fusion} dataset  comprised 492 unique fingervein identities captured in two sessions with six samples each. Thus, the proposed method (and the existing methods employed in this work that includes multiple attention \cite{ma2023fingerMultiAtten} and deep fusion \cite{boucherit2022fingerFUSION}) are trained on the FV-300 dataset and fine-tuned using the first session data (from FV-USM dataset)  that has 6 samples per subject. Testing was performed using the second-session data (from FV-USM dataset) with six samples per subject.  However, the conventional fingervein state-of-the-art techniques (MCP \cite{miura2007extraction}, RLT \cite{RepeatedlineTracking} and WLD \cite{FVPen1}) employed in this study do not require a training set for learning. Therefore, we used the first-session data from the FV-USM dataset as enrolment, and the second session data were used for testing.  This resulted in 492 $\times$ 6 = 2952 genuine and 492 $\times$ 491 $\times$ 6 = 1449432 impostor scores. 

The verification performance of the fingervein algorithms (deep learning based on the proposed method) on the FV-PolyU dataset was performed using a procedure similar to that discussed for the FV-USM dataset. The fingervein algorithms trained on the FV-300 dataset were fine-tuned using  the FV-PolyU dataset. The FV-PolyU dataset \cite{FP_FV_Sensor2} employed in this work  comprises 156 unique identities, from which the finger vein index and middle fingers are captured in two sessions with six samples each. Thus, the FV-PolyU dataset has 312 unique identities, and data from the first session are used to fine-tune both the proposed and SOTA deep learning methods,  which include multiple attention \cite{ma2023fingerMultiAtten} and deep fusion \cite{boucherit2022fingerFUSION}) that are trained on the FV-300 dataset. Testing was performed on the  second session data, which resulted in 312 $\times$ 6 = 1872 genuine and 312 $\times$ 311 $\times$ 6 = 582192 impostor scores.

\begin{table*}[htp]
\centering
\caption{Verification performance of the proposed method with normal and enhanced fingervein data}
\label{tab:Indi}
\resizebox{1.5\columnwidth}{!}{%
\begin{tabular}{|l|l|l|lll|}
\hline
\rowcolor[HTML]{ECF4FF} 
\cellcolor[HTML]{ECF4FF} & \cellcolor[HTML]{ECF4FF} & \cellcolor[HTML]{ECF4FF} & \multicolumn{3}{l|}{\cellcolor[HTML]{ECF4FF}\textbf{TAR = (100-FNMR\%) @ FMR =}} \\ \cline{4-6} 
\rowcolor[HTML]{ECF4FF} 
\multirow{-2}{*}{\cellcolor[HTML]{ECF4FF}\textbf{Data Type}} &
  \multirow{-2}{*}{\cellcolor[HTML]{ECF4FF}\textbf{Algorithms}} &
  \multirow{-2}{*}{\cellcolor[HTML]{ECF4FF}\textbf{EER(\%)}} &
  \multicolumn{1}{l|}{\cellcolor[HTML]{ECF4FF}\textbf{1\%}} &
  \multicolumn{1}{l|}{\cellcolor[HTML]{ECF4FF}\textbf{0.1\%}} &
  \textbf{0.01\%} \\ \hline
Normal Fingervein        & Proposed Method          & 1.85                     & \multicolumn{1}{l|}{97.89}       & \multicolumn{1}{l|}{83.55}       & 54.18      \\ \hline
Enhanced   Fingervein    & Proposed Method          & 1.13                     & \multicolumn{1}{l|}{98.87}       & \multicolumn{1}{l|}{90.53}       & 60.76      \\ \hline
\end{tabular}%
}
\end{table*}

\subsection{Results and discussion}
Table \ref{tab:Results} shows the quantitative performance of the proposed and existing fingervein verification techniques on both FV-300, FV-PolyU and FV-USM datasets, and Figure \ref{fig:DET} shows the DET curves. Existing methods were trained using enhanced fingervein images to optimise the best performance.  Based on the results, the following are important observations: 

\begin{itemize}
\item Training and testing on the same dataset will indicate the improved verification results of the deep learning  based techniques. Therefore, the performance of the deep learning techniques   indicated an improved performance on FV-300 compared to FV-USM and FV-PolyU dataset. 
\item Traditional fingervein techniques (MCP \cite{miura2007extraction}, RLT \cite{RepeatedlineTracking} and WLD \cite{FVPen1})  that are based on template matching using correlation  indicates the superior  performance on the FV-300 dataset compared to FV-USM and FV-PolyU dataset. However, the performance of  RLT \cite{RepeatedlineTracking} and WLD \cite{FVPen1} do not indicate a significant difference in the verification performance between three different fingervein datasets employed in this work. 
\item  Among three traditional fingervein techniques employed in this work, the  MCP \cite{miura2007extraction} indicated the best performance on both datasets. Furthermore, MCP \cite{miura2007extraction} demonstrated improved performance compared to the state-of-the-art deep learning methods employed in this study. 
\item The proposed method has indicated an outstanding verification performance with EER = 0.54\% and TAR = 90.36\% @ FMR = 0.01\% on FV-300 dataset. The proposed method also indicated the best performance with an EER of 15.35\% on the FV-USM dataset. Similar performance is also noted on the FV-PolyU dataset with an EER = 5.52\% . However, the verification performance degraded at a lower FMR on both FV-USM and FV-PolyU dataset. 
\item Based on the results, it is worth nothing that, the deep learning techniques depends on the training data and indicate the limitation to generalize on the another dataset due to the limited number of samples available for fine-tuning. However, compared with existing deep learning methods, the proposed VeinAtnNet exhibits superior verification performance on three fingervein datasets employed in this work.  

\end{itemize}
\subsection{Ablation Study of the proposed method}

\begin{table}[]
\centering
\caption{Ablation study of the proposed method on FV-300 dataset}
\label{tab:Ablation}
\resizebox{\columnwidth}{!}{%
\begin{tabular}{|ccc|c|c|}
\hline
\multicolumn{3}{|c|}{Consecutive Convolution Layers (CCL)} & \multirow{2}{*}{\begin{tabular}[c]{@{}c@{}}Multi-head \\ Self-Attention (MSA)\end{tabular}} & Proposed method \\ \cline{1-3} \cline{5-5} 
\multicolumn{1}{|c|}{Conv-1} & \multicolumn{1}{c|}{Conv-2} & Conv-3 &  & EER (\%) \\ \hline
\multicolumn{1}{|c|}{\cmark}       & \multicolumn{1}{c|}{X}      & X      &\cmark  & 8.29     \\ \hline
\multicolumn{1}{|c|}{\cmark}       & \multicolumn{1}{c|}{\cmark}       & X      & \cmark & 2.38     \\ \hline
\multicolumn{1}{|c|}{\cmark}       & \multicolumn{1}{c|}{\cmark}       &  \cmark      & \cmark & 0.54     \\ \hline
\end{tabular}%
}
\end{table}

\begin{figure}[htp]
\begin{center}
\includegraphics[width=0.9\linewidth]{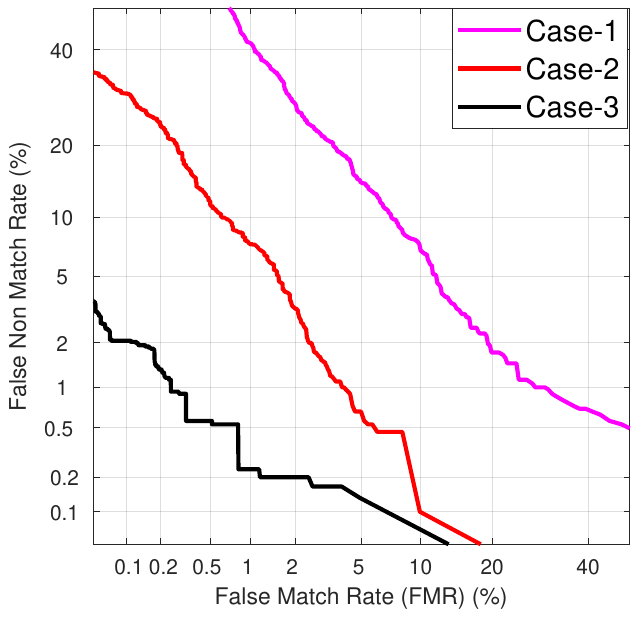}
\end{center}
   \caption{DET Curves indicating the performance of the proposed method with different cases of ablation study}
\label{fig:ablation}
\end{figure}
In this section, we present an ablation study of the proposed method by using the FV-300 dataset. We considered three different cases in which Case-1 represent the performance with Conv-1 and MSA together. Case-2 shows the performance of Conv-1, Conv-2, and MSA while Case-3 indicates the performance of the proposed method with Conv-1, Conv-2, Conv-3, and MSA. Table \ref{tab:Ablation} and Figure \ref{fig:ablation} show the performance of the proposed method for different ablation studies.  The addition of convolutional layers with MSA can improve the overall performance of the proposed VeinAtnNet for reliable fingervein verification.  
\begin{figure}[htp]
\begin{center}
\includegraphics[width=.9\linewidth]{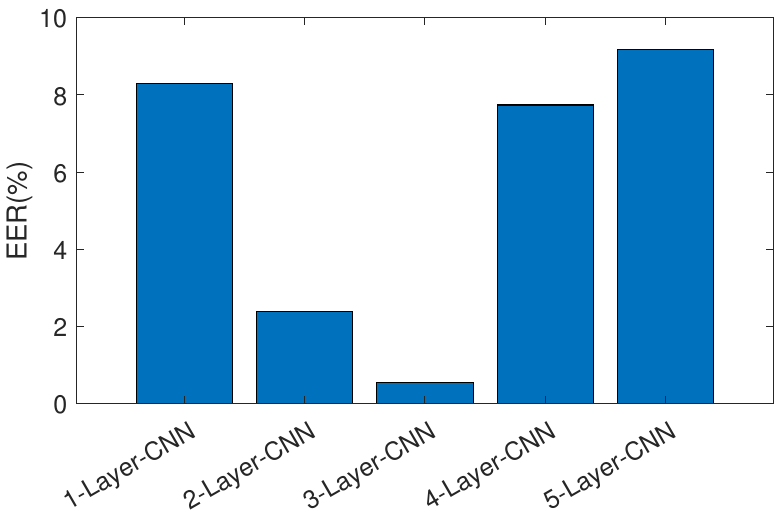}
\end{center}
   \caption{EER of the proposed method with different number of convolution layers.}
\label{fig:EERAbal}
\end{figure}

We further investigated the role of adding additional convolution layers with MSA to improve the verification accuracy. To this extent, we start computing the verification accuracy starting with one Conv layer and increasing it to five consecutive Conv layers with MSA. Figure \ref{fig:EERAbal} shows the verification performance with EER  for different depths of convolution layers. It should be noted that the use of three consecutive layers with MSA can achieve the best performance  and further increase the depth by adding convolution layers. This further justifies the choices made in designing the proposed  method that has indicated the best generalized verification performance compared to the five different SOTA.

\begin{figure}[htp]
\begin{center}
\includegraphics[width=1.0\linewidth]{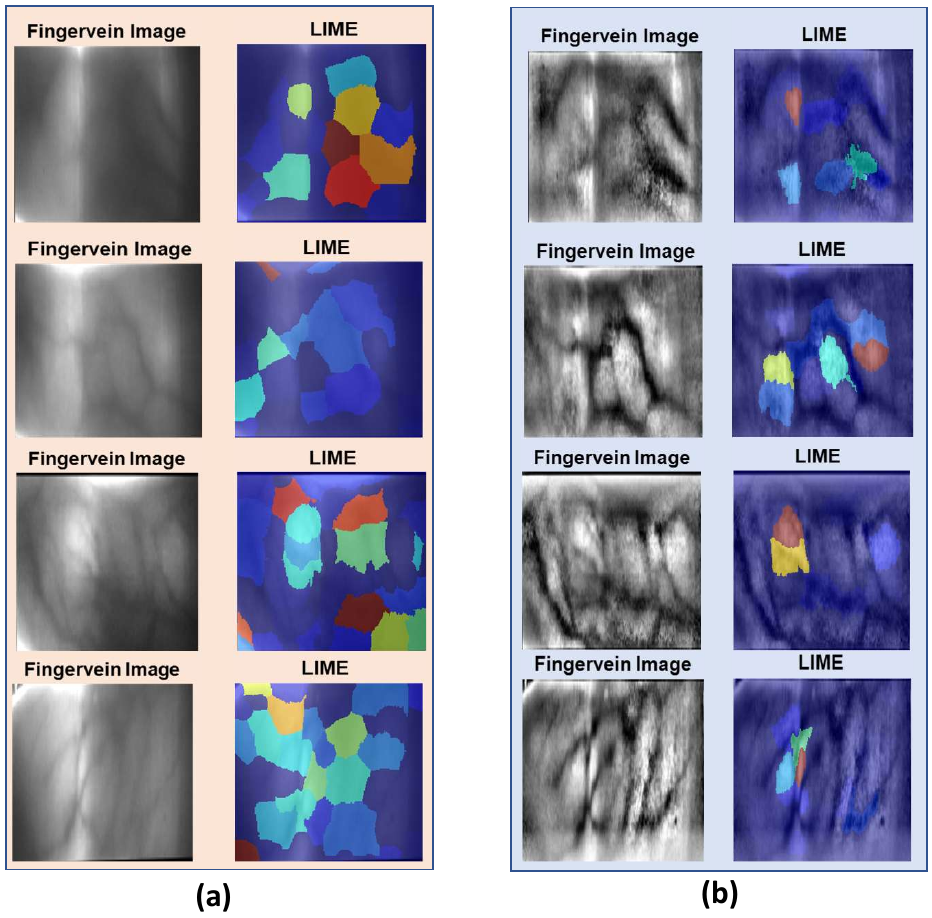}
\end{center}
   \caption{Illustration of LIME based explainability on the proposed method based on the (a) normal and (b) enhanced fingervein images.}
\label{fig:explain}
\end{figure}
\subsection{Interpretation of the proposed method}
To interpret the decision achieved by the proposed method, we employed Local interpretable model-agnostic explanations (LIME) \cite{ribeiro2016shouldlime} to explain the perdition's on the probe fingervein images. Because the proposed method is based on both normal and enhanced fingervein images, we present the qualitative and quantitative results for both fingervein image types. Table \ref{tab:Indi} indicates the quantitative performance of the proposed method with normal and enhanced image alone. The obtained results indicated a similar verification performance with EER and higher FMR values. However, with lower FMR values, the proposed method exhibited better performance with enhanced fingervein samples. Thus, the availability of the enhanced fingervein pattern indicates more discriminant information to improve verification accuracy at low FMR values.

Figure \ref{fig:explain} shows the qualitative results of the LIME method for visualizing important regions in the fingervein image, which has contributed to successful verification. The LIME explanation is shown on the fingervein images from the FV-300 dataset for successful verification prediction at FAR = 0.01\%. As shown in Figure \ref{fig:explain}, the proposed method utilises  more image regions with normal fingervein images compared with the enhanced  fingervein to make the decision. However, with enhanced fingervein images, the decision is based on a smaller number of regions associated with vein pattern and particularly on the minutiae points of the fingervein. These observations justify the improved performance of the proposed method with enhanced fingervein images compared with normal fingervein images.  

\section{Conclusion}
\label{sec:conc}
Fingervein biometrics are widely employed in various secure access control applications. In this study, we proposed a novel method based on a convolutional multihead attention module for reliable fingervein verification.  The proposed VeinAtnNet is based on three consecutive convolution layers and multihead attention with four heads and 64 channels connected in parallel to the normal and enhanced fingervein samples. Finally, the decision is made using the score-level fusion of the normal and enhanced fingerveins. Extensive experiments were performed on both publicly and newly collected finger vein datasets. The quantitative performance of the proposed method was benchmarked using five state-of-the-art fingervein verification methods. The obtained results indicate the superior performance of the proposed method on both publicly available and newly collected fingervein datasets. 
{\small
\bibliographystyle{ieee_fullname}
\bibliography{VeinSensor}
}

\end{document}